\documentclass[10pt,twocolumn,letterpaper]{article}

\usepackage{iccv}
\usepackage{times}
\usepackage{epsfig}
\usepackage{graphicx}
\usepackage{amsmath}
\usepackage{amssymb}
\usepackage{booktabs}
\usepackage{multirow}
\usepackage{makecell}
\usepackage{float}
\usepackage[accsupp]{axessibility}
\usepackage{dsfont}

\usepackage{graphicx} 
\usepackage{caption}
\usepackage{subcaption}
\makeatletter
\@namedef{ver@everyshi.sty}{}
\makeatother
\usepackage{tikz}
\usepackage{pgf}

\def\Ours{PoseMatcher}
\usepackage[pagebackref=true,breaklinks=true,letterpaper=true,colorlinks,bookmarks=false]{hyperref}

\iccvfinalcopy 

\def\iccvPaperID{5543} 

\ificcvfinal\fi

\begin{document}

\title{PoseMatcher: One-shot 6D Object Pose Estimation by Deep Feature Matching}

\author{Pedro Castro\\
Imperial College London\\
{\tt\small p.castro18@imperial.ac.uk}
\and
Tae-Kyun Kim\\
Imperial College London, KAIST\\
{\tt\small tk.kim@imperial.ac.uk}
}

\maketitle
\ificcvfinal\thispagestyle{empty}\fi


\begin{abstract}

Estimating the pose of an unseen object is the goal of the challenging one-shot pose estimation task. Previous methods have heavily relied on feature matching with great success. However, these methods are often inefficient and limited by their reliance on pre-trained models that have not be designed specifically for pose estimation. In this paper we propose \textbf{PoseMatcher}, an accurate model free one-shot object pose estimator that overcomes these limitations. 
We create a new training pipeline for object to image matching based on a three-view system: a query with a positive and negative templates. This simple yet effective approach emulates test time scenarios by cheaply constructing an approximation of the full object point cloud during training.
To enable PoseMatcher to attend to distinct input modalities, an image and a pointcloud, we introduce IO-Layer, a new attention layer that efficiently accommodates self and cross attention between the inputs.
Moreover, we propose a pruning strategy where we iteratively remove redundant regions of the target object to further reduce the complexity and noise of the network while maintaining accuracy.
Finally we redesign commonly used pose refinement strategies, zoom and 2D offset refinements, and adapt them to the one-shot paradigm.
We outperform all prior one-shot pose estimation methods on the Linemod and YCB-V datasets as well achieve results rivaling recent instance-level methods. The source code and models are available at \href{https://github.com/PedroCastro/PoseMatcher}{github.com/PedroCastro/PoseMatcher}.


\end{abstract}


\section{Introduction}

Accurately retrieving the relative position and orientation of an object is the first step for any task that requires interaction with objects in the real world. Estimating the pose of an object is an indispensable step for robotic manipulation as well as in VR/AR applications. It is imperative for the pose estimation to be accurate and robust to external obstacles such as occlusion, illumination and symmetries. Existing methods excel at retrieving the pose of known objects to a very high accuracy standard \cite{gdr, sopose, SC6D}. Some methods can even produce pose estimation at high throughput \cite{gdr, crt6d}, be specially robust to symmetries \cite{surfemb} and can even surpass synthetic to real domain gaps \cite{self6d, juil}. However, most share a very large limitation: the target objects must be known. This constraint is severely limiting as it necessitates model retraining for each object addition. Retraining new objects on existing models might lead to catastrophic forgetting if not handled properly~\cite{catastrophic}. Category level approaches~\cite{posefromshape, nocs} try to generalize up to a category, where each target object can be obtained with a simple deformation of a canonical model and all share semantic keypoints (ex. handle on a mug or the cap of a bottle). Nonetheless, much of the same one-shot domain problems still remain: objects category must have been seen before and the target object must not lie outside of the scope of the category for the estimation to be possible. In order to overcome these problems we must tackle object pose estimation in the one-shot paradigm.

\begin{figure}[tb]
  \resizebox{\linewidth}{!}{\includegraphics[]{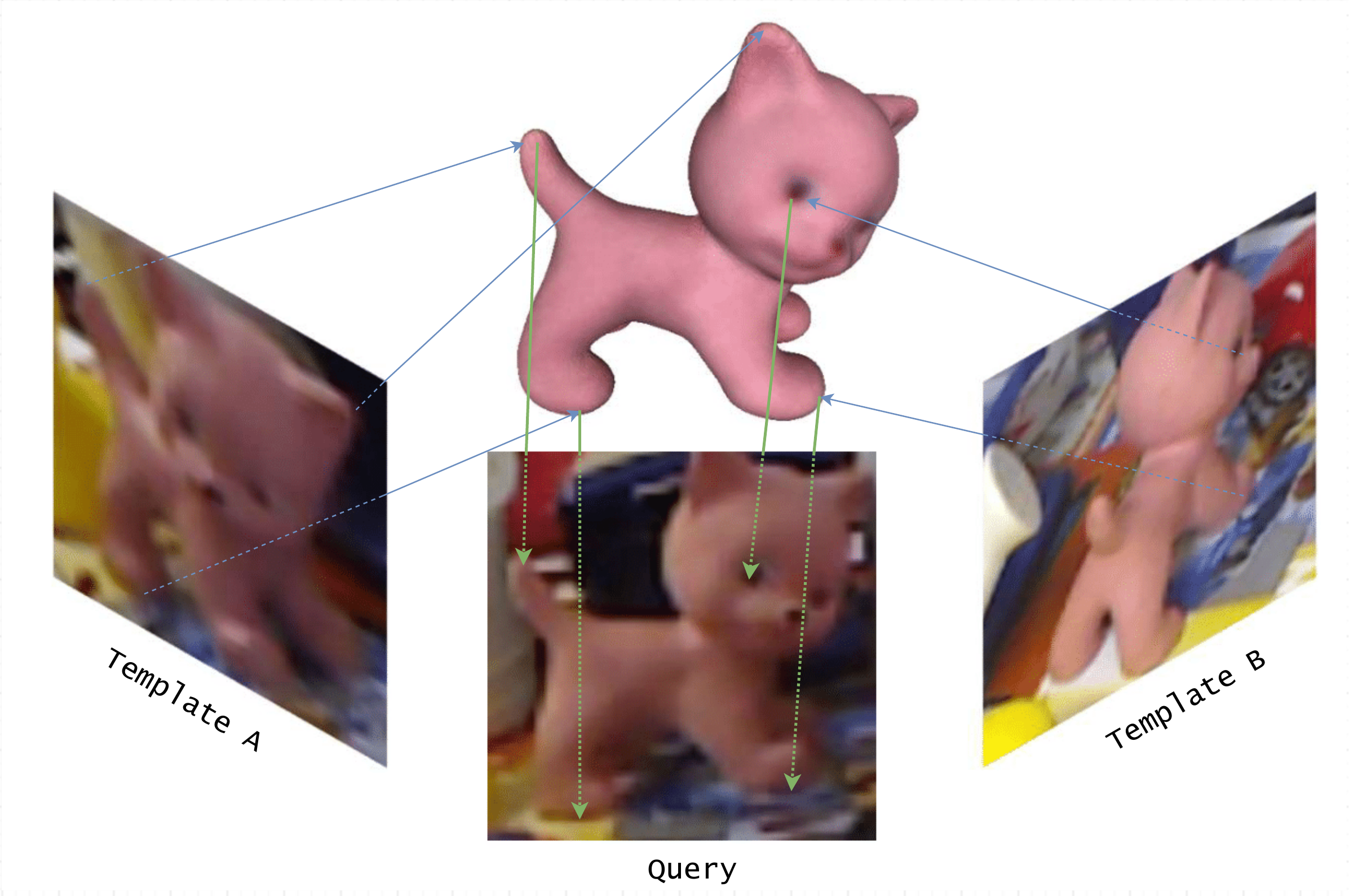}}
  \label{fig:training pipeline}
  \vspace{-20pt}
\caption{\textbf{Illustrative diagram of PoseMatcher training diagram.} For each training instance query we sample two templates. From these templates, we reconstruct a partial point cloud simulating the object point cloud at test time.}
\vspace{-10pt}
\label{fig:init_image}
\vspace{-10pt}
\end{figure}

By one-shot pose estimation, we refer to estimating the pose of an novel object based on information seen only at test-time, without the object,or its category, being present in the training dataset. Older methods have some very hard constraints such as needing masking and depth maps at test-time \cite{ppf}, a full colored 3D model of the target object \cite{deepim}, need to be fine-tuned on images of the same dataset for difficulties of overcoming domain gap~\cite{gen6d} or rely on an extremely expensive test-time optimization through renderers~\cite{inerf, nerfpose, latentfusion}.
More recently, feature matching methods have been shown to achieve impressive results. Particularly, OnePose~\cite{onepose} and OnePose++~\cite{onepose++} make use of existing state of the art pre-trained descriptor extractors on top of which a pose estimation pipeline is built. However, by relying on a fixed pre-trained model to extract descriptive keypoints from templates it fails to capture the optimal keypoint object descriptions. Ideally, the template extraction model should be the same as the query model and should be jointly optimized.

We rethink the approach to the problem and introduce a three-view pipeline that allows us to jointly train the template and query extractors. Just by redesigning the training pipeline, we can improve OnePose++\cite{onepose++} without additional changes to its methodology. We also introduce a new efficient image to object attention layer which we call IO-Layer, which reduces both parameters  when compared to the modules used by OnePose++~\cite{onepose++} and also separates the two input modalities, an image and a pointcloud, allowing the model to optimize weights specifically for either image or pointcloud keypoints. We also found that working with a full object at all stages of the feature matching is redundant and reduces pose estimation accuracy which lead us to introduce a novel iterative matching based pointcloud prunning. On top of these changes, we employ a 3D-refinement technique based on zoom refinement used on instance-level pose estimation~\cite{deepim,crt6d,cosypose}.

In summary our contributions are as follows:

\begin{itemize}
    \item We redesign the training pipeline for one-shot object-to-image matching. Our three-view training approach allows us to train from scratch PoseMatcher, a pose estimation method leveraging detection free keypoint matching.
    \item We introduce a more efficient image to object attention layer we call IO-Layer, specifically built to accommodate the two input modalities, the image and the object pointcloud. 
    \item We propose a layered pruning of the target object point cloud. We improve runtime by reducing the amount of \textit{attented} keypoints while reducing noisy matching, leading to an improvement in accuracy.
    \item We create a fine level 3D based refinement where we directly estimate relative 2D and depth positions, replacing the 2D keypoint refinement used by prior one-shot approaches.
\end{itemize}

\section{Literature}

\begin{figure*}[tb]
  \resizebox{\linewidth}{!}{\includegraphics[]{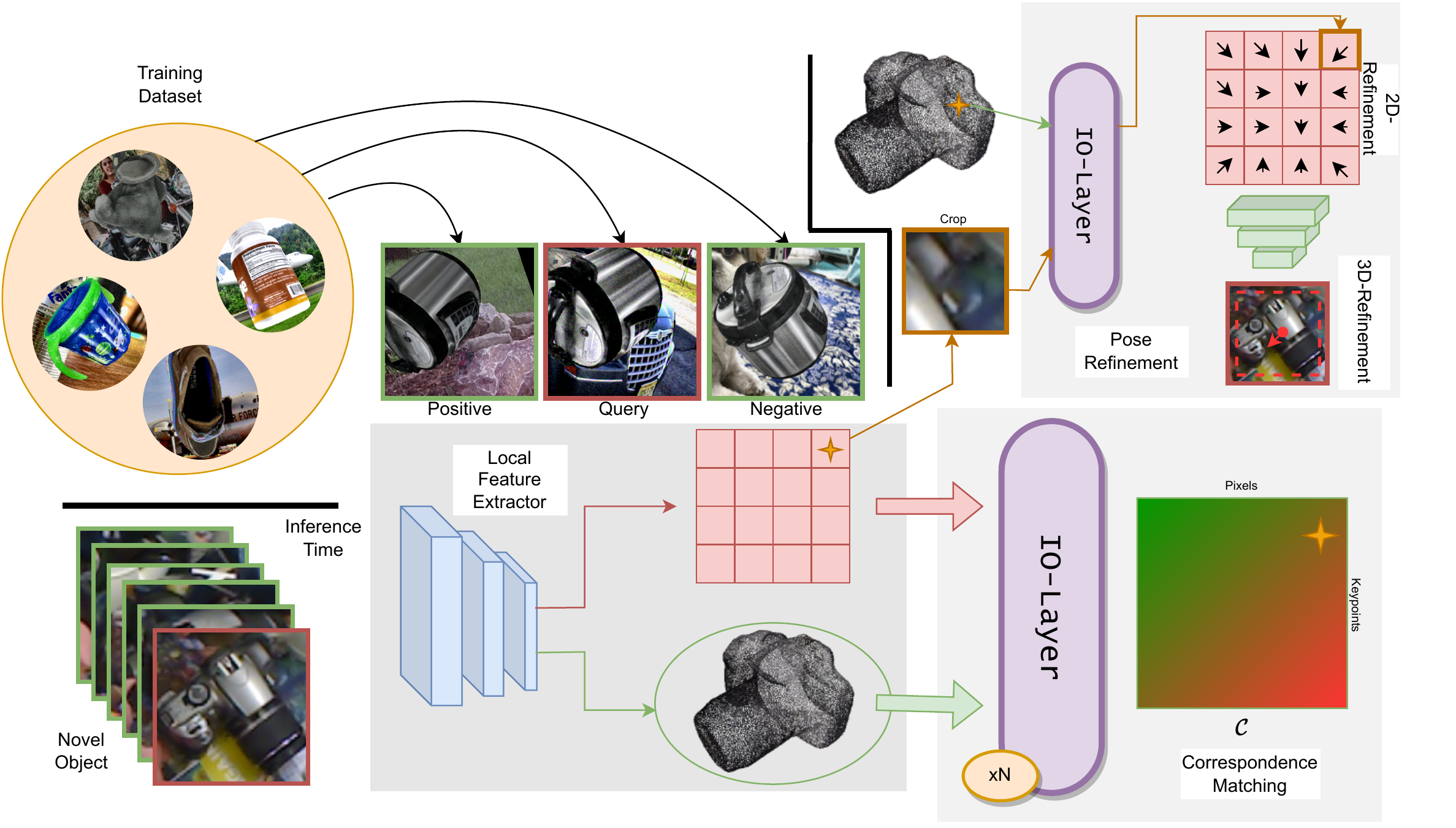}}
  \label{fig:training pipeline}
  \vspace{-10pt}
\caption{\textbf{Diagram of PoseMatcher.} At training time we sample from the Google Scanned Objects\cite{goog_objects} dataset and from two opposing views templates we construct a partial point cloud. At test time, a novel unseen object is used instead. We extract local features from both the query and the templates. Using our novel IO-Layer, we compute dense correspondences between the image and the object. We then take the matches (yellow star) and further refine them using 2D and 3D techniques.}
\label{fig:init_image}
\vspace{-10pt}
\end{figure*}

\noindent\textbf{Fully supervised pose estimation.} 
Instance-level pose estimation is designed to support a single object. Due to this task's narrowed scope, the accuracy of instance-level methods are becoming increasingly more impressive. 
Early 6D pose estimation works focused on recovering the 2D position of specific keypoints, mainly the 3D bounding box of the target~\cite{bb8, heatmaps,seamless}. PVNet~\cite{pvnet} found that choosing keypoints that lie within the object's silhouette would yield better results.
This idea has become a mainstay among current keypoint based methods \cite{seg_driven,crt6d}. An alternative to chosing a limited amount of features was to estimate the coordinate of the surface of the object at each pixel \cite{nocs, dpod, pix2pose}. GDR\cite{gdr} and SO-Pose\cite{sopose} approximated PnP through a small neural network and optimizing directly through pose errors. Direct pose estimation with a combination of a dense intermediate representation makes up most of the state of the art methods as per the BOP challenge~\cite{bop}. 
Other ideas have focused on designing textures as learnable features that can be more easily estimated \cite{surfemb, repose} and as a result, better correspondences are found.
 In order to decrease the need for real data, Sock \etal \cite{ juil} proposed a differentiable pipeline where learning is done through comparison with a rendering of the estimation which is done using keypoints, where Self6D \cite{self6d} directly outputs pose.

\noindent\textbf{One-shot pose estimation.}
Until recent versions of the BOP challenge~\cite{bop}, Point Pair Features~\cite{ppf} held the top spot on its leaderboard. It relies on selecting geometrical relevant keypoints from an existing 3D and matching these to a depth map. This method relies on capturing information with a depth sensor at inference time which is not commonly available. 
Extending the prior instance-level approaches to category-level allows for object deformation and texture change within a narrow category \cite{wild6d2, 6dwild, nocs}.
The main idea from NOCS \cite{nocs}, although used for instance-level by a wide range of works \cite{gdr, sopose, dpod}, was initially to predict pose of slightly deformable objects within a category, as it was inspired by its human reconstruction contemporary DenseBody~\cite{densebody} . Recent advancements have been made that allow for training for in the wild \cite{6dwild}. However, these are sill limited by the similarity to objects of the same category. Our aim is to overcome this limitation. 

Methods such as Pitteri~\etal\cite{unseen1} and CorNet\cite{unseen2} rely on training with a subset of similar objects of the same dataset and/or scene, which includes biases towards illuminations, noise, background and shape. 
Recent attempts have shown researchers are keen on tackling one-shot pose estimation. OSOP~\cite{osop} introduces a global template matching method where a given query is matched to the closest viewpoint from a database of pre-processed synthetic viewpoints, which requires the object model, an assumption we do not make. Gen6D \cite{gen6d} works in a similar fashion by matching a query to a small amount of real annotated images and then refining the pose. However, it is susceptible to poor pose initialization and poor 2D bounding box detections.
The closest work to our own is OnePose++, an improved version of OnePose\cite{onepose}, where the feature extraction is replaced by the detector free feature matcher LoFTR~\cite{loftr}. However, OnePose++\cite{onepose++} adapts a pre-trained LoFTR to one-shot pose estimation without taking consideration the different modalities of image and point cloud, self-occlusion redundancy and the possibility of using 6D pose based refinement.

\section{PoseMatcher}

The goal of the PoseMatcher is to establish matches between the two sets of keypoints. We establish correspondence between the 2D keypoints features extracted from the query image $\mathcal{I}^Q$ and the object point cloud $\{\mathcal{P}_O\}$. From those matches we can apply PnP and recover full 6D pose $\zeta$. A diagram of PoseMatcher can be seen in Fig.~\ref{fig:init_image}.

\subsection{Template Based Training}

We adopt a new training methodology that allows us to design a one-shot object-image detector free feature matching model from scratch, specifically aimed at pose estimation task. We therefore remove the need for pre-trained descriptor extractors used by OnePose and OnePose++ \cite{onepose, onepose++}.

However, a single viewpoint will result in a partial reconstruction of the object. At inference time, both the visible and occluded regions of the object will be represented in the template point-cloud. Therefore, a single template is not enough to emulate the conditions at test-time.

To address these limitations, we sample an additional template image we refer to as the negative template $\mathcal{I}^-$. This template shares low co-visibility area with the anchor image. The keypoints extracted from $\mathcal{I}^-$ serve to generate nearly complete reconstruction of the target object, with the visible sections being sampled from $\mathcal{I}^+$ while the self-occluded ones from $\mathcal{I}^-$. This 2-template paradigm simulates the full object point-cloud available at the inference time.

In order to optimize through the matching task we apply the differentiable dual-softmax operator as proposed by LoFTR~\cite{loftr} and subsequently used by both OnePose and OnePose++~\cite{onepose,onepose++}. 

In order to output matches, we start by using a local feature extractor, such as a Resnet~\cite{resnet} to extract coarse and fine level feature maps, $\hat{\mathcal{F}}_{\mathcal{I}^Q} \in \mathbb{R}^{HxWx\hat{C}}$ and $\Tilde{\mathcal{F}}_{\mathcal{I}^Q} \in \mathbb{R}^{HxWx\Tilde{C}}$ from the query image. We repeat the same process for both $\mathcal{I}^+$ and $\mathcal{I}^-$ and sample $M$ points from both templates. We make sure that the template positions lie within the objects mask such that:
\begin{equation}
\begin{split}
    p^+ = \{p_k^+|~k \in \mathcal{M}^+ \} , \\
    p^- = \{p_k^-|~k \in \mathcal{M}^- \} , 
\end{split}
\end{equation}

where $\mathcal{M}$ refers to the segmentation mask. We backproject, using the a depth map only available at training time, and generate a template point ${\mathcal{P_{\mathcal{T}}}^{3D}} \in \mathbb{R}^{2Mx3}$ with its corresponding extracted features ${\hat{\mathcal{F}_\mathcal{T}}} \in \mathbb{R}^{2Mx\hat{C}}$ and ${\Tilde{\mathcal{F}_\mathcal{T}}} \in \mathbb{R}^{2Mx\Tilde{C}}$, where $\mathcal{T}$ refers to the combined positive and negative templates.

At this point, we perform dense matching between the coarse level query features $\hat{\mathcal{F}}_{\mathcal{I}^Q}$ and the extracted point cloud ${\hat{\mathcal{F}_\mathcal{T}}}$. We now aim to globally match each pixel of the query image that contains the object to its respective point cloud keypoint. We make use of positional embeddings to encode positional information into each feature point. For 2D keypoints we use
the sinusoidal fixed version used by DETR~\cite{detr} while for 3D positional encoding we use a simple 3-layer MLP as used by SurfEmb~\cite{surfemb} and OnePose++~\cite{onepose++}.

We flatten both feature maps and apply self and cross attention layers following other feature matching papers~\cite{superglue, loftr, onepose, onepose++, matchformer} to generate more easily separable features on each set. Our objective is to construct a matrix $\mathds{P}^{HWx2M}$ that reflects the level of confidence of the correspondence between $\mathcal{P}^{2D}$ and $\mathcal{P}^{3D}$. A score matrix $\mathcal{S}$ is computed by measuring the cosine similarity between the two sets of transformed features in a contrastive manner. We apply dual-softmax \cite{dualsoftmax} over $\mathcal{S}$ to calculate the correspondence confidence matrix $\mathds{P}$:

\vspace{-10pt}
\begin{equation}
    \mathds{P} = softmax(\mathcal{S}(i,\cdot))_c~\cdot~softmax(\mathcal{S}(\cdot,c))_i,
\end{equation}

\begin{figure*}[]
  \vspace{-15pt}
  \resizebox{\linewidth}{!}{\includegraphics[]{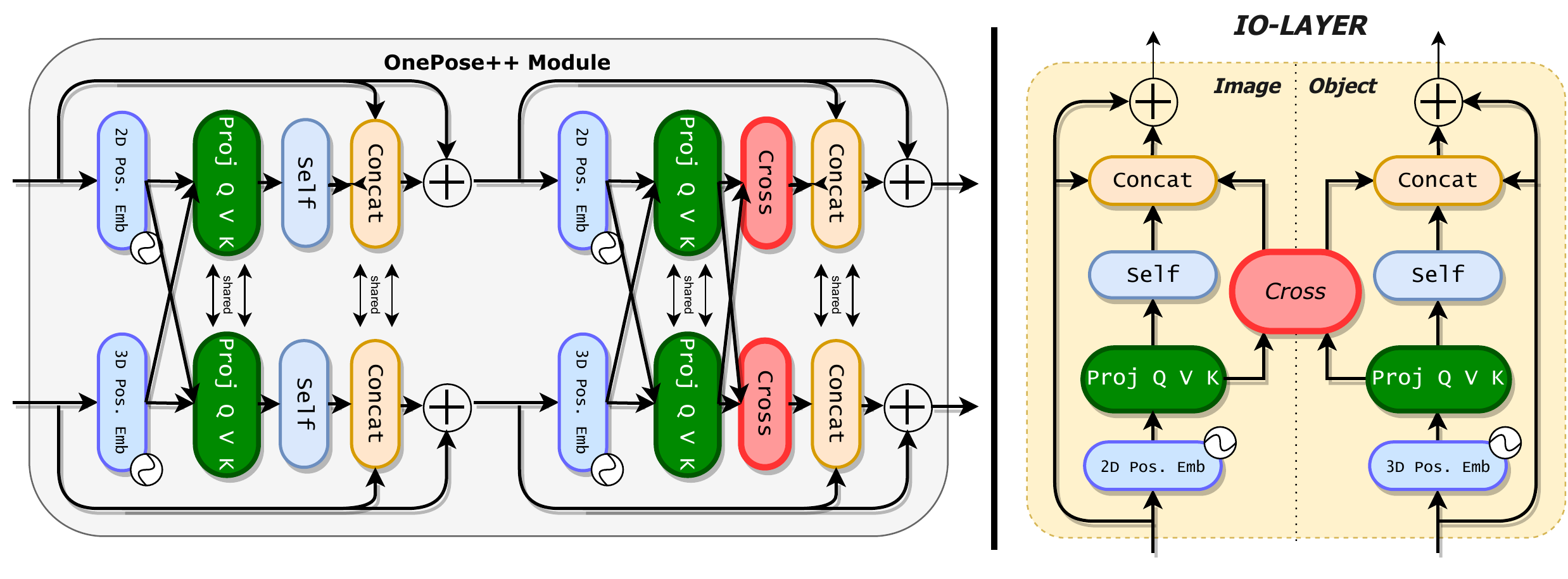}}
  \label{fig:iolayer}
  \vspace{-20pt}
\caption{\textbf{IO-Layer architecture.} We improve on the attention modules used by LoFTR and OnePose++ \cite{loftr, onepose}. We have modality specific projections for image features embedded with 2D positional encoding and for object template features with its 3D positional encoding. We found that separating the two modalities improves the final pose accuracy.}
  \vspace{-15pt}
\label{fig:iolayer}
\end{figure*}

$\tau$ being a temperature hyperparameter, with $i$ and $c$ being the indices of the flattened image pixels and point cloud, respectively. The image to object correspondences $\mathcal{C}$ are established by choosing only those that exceed a certain level of confidence, represented by the threshold value $\theta$, and meet the mutual nearest neighbor (MNN) constraint to remove false matches:

\vspace{-10pt}
\begin{equation}\label{eq:matching}
    \mathcal{C} = \{ (i , c) | \forall (i, c) \in MNN(\mathcal{P}_{i}^{2D}, \mathcal{P}_{c}^{3D}), \mathds{P}_{i, c} \ge \theta \}.
\end{equation}

The coarse loss $\mathcal{L}_c$ used to optimize $\mathcal{C}$ uses focal loss~\cite{focalloss} as suggested by LoFTR~\cite{loftr}. 

\subsection{IO-Layer: An object-image attention layer}

Image to image feature matching has seen its effectiveness increase with the introduction of self and cross attention mechanisms \cite{matchformer, superglue, loftr, paraformer}. These layers are particularly efficient for this task seeing as both inputs are from the same modality and share the same spatial structure. For this reason, self attention layers can share weights and cross attention only requires swapping the query and template image, as you can see in Fig~\ref{fig:iolayer}. OnePose++~\cite{onepose++} has suggested using the same approach for image to object matching however in this scenario we are working with two different modalities, a 2D image and a 3D point cloud. 

We postulate that mixing modalities has an adverse effect on learning distinctive features. The positional encoding for each input is structurally different which means that attention layers will have difficulty separating the features of each modality. OnePose++~\cite{onepose++} shares a sign of this problem when they observe that adding 3D positional encoding results in a small ($<1\%$) improvement.
A simple ad-hoc solution would be to have modality specific weights, however that would double the amount of parameters and subsequently the necessary amount of training data and compute.
In standard self attention layers an input sequence is linearly projected into a query, key and value: 
\textbf{$Q_i$},\textbf{$K_i$},\textbf{$V_i$},stemming from the same input $i$. 
For cross attention, one would linearly project  \textbf{$Q_i$}, \textbf{$K_j$}, \textbf{$V_j$} where $i$ and $j$ are different inputs. Usually, in attention mechanisms, the encoded message is parsed to the decoder and is not returned. However, since it is important for the image features to be distinct from the object's and vice-versa, the message must be passed bi-directionally and therefore requires careful redesign.

We propose a simple layer, specially designed for \textbf{I}mage to \textbf{O}bject matching we call \textbf{IO-Layer}. (\textbf{$Q_i$},\textbf{$K_i$},\textbf{$V_i$}) and (\textbf{$Q_o$},\textbf{$K_o$},\textbf{$V_o$}), the inputs to the attention mechanism for inputs and object respectively, are computed only once and used for both self and cross attention. Therefore cross attention is computed as $softmax(Q_iK_o^T/\sqrt{d})K_o$ and $softmax(Q_oK_i^T/\sqrt{d})K_i$ for image to object and object to image attention respectively, where we only have to project each sequence once. We modify self and cross attention accordingly to support Linear Attention~\cite{linear_attention}.
Much like the layers used in LoFTR and OnePose++, these can be stacked together. We exemplify the IO-Layer on Fig.\ref{fig:iolayer}.

\subsection{Object Pruning}

OnePose++ proposes sampling and using a point cloud template with over 15k keypoints per object. Performing attention over such a higher number of keypoints is expensive even if using more efficient mechanisms such as Linear Attention~\cite{linear_attention}.
The set of keypoints that are not in visible in the query image should be quickly identifiable in early matching. Removing these keypoints from the point cloud allows for better separability of the features of the remaining visible keypoint, which leads to a better match.  As a peripheral advantage, by removing a set of keypoints from the template point cloud, we reduce the complexity of the following attention layers. During pruning, we do not impose a hard threshold on the confidence of the matches but rather but rather we select the keypoints with higher confidences of existing in the image. We extend the matching operation in Eq.~\ref{eq:matching} where we sum  the keypoint confidences over every pixel and prune the lowest ones.
We insert a pruning step after each IO-Layer. The amount of pruning is subjected to ablation studies in Section~\ref{sec:experiments}.

\subsection{Pose Refinement}

We add a fine level 2D-refinement post-processing step as described by LoFTR~\cite{loftr} and OnePose++\cite{onepose++}. For each match, we crop a window at the match location and perform simple matching w.r.t. the matched keypoints. We supervise it by minimizing the Euclidean distance between the center of the grid and the projected keypoints. More details about 2D refinement are available in LoFTR~\cite{loftr}. 

However, its impact is limited as we show in the experimental section. We believe this is due to the use of PnP following the keypoint position refinement. Prior 6D object pose literature has shown that estimating translation through PnP leads to poor results \cite{cdpn, gdr}. An alternative approach, first introduced by CDPN \cite{cdpn}, is to directly estimate the translation as the 2D offset of the object centroid as well as bounding box relative depth estimation. Directly applying this step to PoseMatcher cannot be done seeing as the objects are not known. 

In order to perform translation estimation, we adopt translation refinement, which will refer to as 3D-refinement, a strategy used by iterative methods \cite{crt6d, cosypose, deepim}. Instead of estimating the translations directly, we estimate the translation errors given an initial pose. Intuitively, we estimate the necessary 2D translation and \textit{zoom} necessary to align the initial pose to the target pose.

In order to generate the initial pose $\zeta^0$, we perform PnP over the coarse matching keypoints. In the 2D-refinement step, the initial position of the matched keypoints are the image grid default positions. However, if we are performing a refinement over an initial pose, this alignment might be broken due to erroneous matching, pose estimation or even sub-pixel positioning. Therefore we project the matched keypoints using the initial pose and use those 2D locations as the center for the fine-sampling grid crops $F^i_{crop}$. We perform self and cross attention over the grids and the corresponding 3D fine feature $T_f^i$ using a single IO-Layer. We compute the 2D expectation and supervise its output using the same methodology as described for the 2D-refinement.

We propose a lightweight CNN similar to the one used for learned Patch-PnP in GDR-Net~\cite{gdr}. We build our pose representation as a pixel-wise map where at each coarse pixel we collect each matched keypoint refinement prediction (i.e. refinement direction). This CNN outputs the \textit{3D-refinement} in the form of $\Delta T$ and $\Delta Z$, the 2D location and zoom offsets respectively.
Moreover, in order to supervise the refinement step, LoFTR~\cite{loftr} proposes computing the total variance of the matching heatmap in order to penalize more confident but erroneous estimations. We propose reusing this variance to introduce a confidence measure to each keypoint refinement output and append to each input its corresponding refinement.

\begin{table*}[t]
\resizebox{\textwidth}{!}{%
\begin{tabular}{c|cccc|cccccc}
\Xhline{5\arrayrulewidth}    
Type & \multicolumn{3}{c|}{Fully Supervised}                                                                                                                    & \multicolumn{2}{c|}{Self-Supervised}                                   & \multicolumn{4}{c|}{One-Shot}     \\ \hline

Method                   & \multicolumn{1}{c|}{PVNet~\cite{pvnet}} & \multicolumn{1}{c|}{GDR~\cite{gdr}}  & \multicolumn{1}{c|}{SO-Pose~\cite{sopose}}  & \multicolumn{1}{c|}{Self6D~\cite{self6d}} & \multicolumn{1}{c|}{Sock~\etal~\cite{juil}}     & \multicolumn{1}{c|}{Gen6D~\cite{gen6d}} & \multicolumn{1}{c|}{OnePose~\cite{onepose}}  & \multicolumn{1}{c|}{OnePose++~\cite{onepose++}}     & \textbf{PoseMatcher} \\ \hline\hline

Ape                     & \multicolumn{1}{c|}{43.6}  & \multicolumn{1}{c|}{85.9} & \multicolumn{1}{c|}{-}    & \multicolumn{1}{c|}{38.9}    & \multicolumn{1}{c|}{37.6}         & \multicolumn{1}{c|}{-}      & \multicolumn{1}{c|}{11.8}   & \multicolumn{1}{c|}{31.2}          & \textbf{59.2} \\ \hline
Benchvise               & \multicolumn{1}{c|}{99.9}  & \multicolumn{1}{c|}{99.8} & \multicolumn{1}{c|}{-}    & \multicolumn{1}{c|}{75.2}    & \multicolumn{1}{c|}{78.6}         & \multicolumn{1}{c|}{62.1}   & \multicolumn{1}{c|}{92.6}   & \multicolumn{1}{c|}{97.3}          & \textbf{98.1} \\ \hline
Camera                  & \multicolumn{1}{c|}{86.9}  & \multicolumn{1}{c|}{96.5} & \multicolumn{1}{c|}{-}    & \multicolumn{1}{c|}{36.9}    & \multicolumn{1}{c|}{65.6}         & \multicolumn{1}{c|}{45.6}   & \multicolumn{1}{c|}{88.1}   & \multicolumn{1}{c|}{88.0}          & \textbf{93.4} \\ \hline
Can                     & \multicolumn{1}{c|}{95.5}  & \multicolumn{1}{c|}{99.3} & \multicolumn{1}{c|}{-}    & \multicolumn{1}{c|}{65.6}    & \multicolumn{1}{c|}{65.6}         & \multicolumn{1}{c|}{-}      & \multicolumn{1}{c|}{77.2}   & \multicolumn{1}{c|}{89.8}          & \textbf{96.0} \\ \hline
Cat                     & \multicolumn{1}{c|}{79.3}  & \multicolumn{1}{c|}{93.0} & \multicolumn{1}{c|}{-}    & \multicolumn{1}{c|}{57.9}    & \multicolumn{1}{c|}{52.5}         & \multicolumn{1}{c|}{40.9}   & \multicolumn{1}{c|}{47.9}   & \multicolumn{1}{c|}{70.4}          & \textbf{88.0} \\ \hline
Driller                 & \multicolumn{1}{c|}{96.4}  & \multicolumn{1}{c|}{100.} & \multicolumn{1}{c|}{-}    & \multicolumn{1}{c|}{67.0}    & \multicolumn{1}{c|}{48.8}         & \multicolumn{1}{c|}{48.8}   & \multicolumn{1}{c|}{74.5}   & \multicolumn{1}{c|}{92.5}          & \textbf{98.4} \\ \hline
Duck                    & \multicolumn{1}{c|}{52.6}  & \multicolumn{1}{c|}{65.3} & \multicolumn{1}{c|}{-}    & \multicolumn{1}{c|}{19.6}    & \multicolumn{1}{c|}{35.1}         & \multicolumn{1}{c|}{16.2}   & \multicolumn{1}{c|}{34.2}   & \multicolumn{1}{c|}{42.3}          & \textbf{54.1} \\ \hline
Eggbox*                 & \multicolumn{1}{c|}{99.2}  & \multicolumn{1}{c|}{99.9} & \multicolumn{1}{c|}{-}    & \multicolumn{1}{c|}{99.0}    & \multicolumn{1}{c|}{89.2}         & \multicolumn{1}{c|}{-}      & \multicolumn{1}{c|}{71.3}   & \multicolumn{1}{c|}{99.7}          & \textbf{97.8} \\ \hline
Glue*                   & \multicolumn{1}{c|}{95.7}  & \multicolumn{1}{c|}{98.1} & \multicolumn{1}{c|}{-}    & \multicolumn{1}{c|}{94.1}    & \multicolumn{1}{c|}{64.5}         & \multicolumn{1}{c|}{-}      & \multicolumn{1}{c|}{37.5}   & \multicolumn{1}{c|}{48.0}          & \textbf{91.5} \\ \hline
Holepuncher             & \multicolumn{1}{c|}{81.9}  & \multicolumn{1}{c|}{73.4} & \multicolumn{1}{c|}{-}    & \multicolumn{1}{c|}{16.2}    & \multicolumn{1}{c|}{41.5}         & \multicolumn{1}{c|}{-}      & \multicolumn{1}{c|}{54.9}   & \multicolumn{1}{c|}{69.7}          & \textbf{73.4}  \\ \hline
Iron                    & \multicolumn{1}{c|}{98.9}  & \multicolumn{1}{c|}{86.9} & \multicolumn{1}{c|}{-}    & \multicolumn{1}{c|}{77.9}    & \multicolumn{1}{c|}{80.9}         & \multicolumn{1}{c|}{-}      & \multicolumn{1}{c|}{89.2}   & \multicolumn{1}{c|}{97.4}          & \textbf{97.9}  \\ \hline
Lamp                    & \multicolumn{1}{c|}{99.3}  & \multicolumn{1}{c|}{99.6} & \multicolumn{1}{c|}{-}    & \multicolumn{1}{c|}{98.2}    & \multicolumn{1}{c|}{70.7}         & \multicolumn{1}{c|}{-}      & \multicolumn{1}{c|}{87.6}   & \multicolumn{1}{c|}{97.8}          &  \textbf{98.1} \\ \hline
Phone                   & \multicolumn{1}{c|}{92.4}  & \multicolumn{1}{c|}{86.3} & \multicolumn{1}{c|}{-}    & \multicolumn{1}{c|}{50.1}    & \multicolumn{1}{c|}{60.5}         & \multicolumn{1}{c|}{-}      & \multicolumn{1}{c|}{60.6}   & \multicolumn{1}{c|}{76.0}          &  \textbf{92.1}\\ \hline
\hline
Average                 & \multicolumn{1}{c|}{86.3}  & \multicolumn{1}{c|}{91.0} & \multicolumn{1}{c|}{96.0}    & \multicolumn{1}{c|}{58.9} & \multicolumn{1}{c|}{60.6}         & \multicolumn{1}{c|}{-}      & \multicolumn{1}{c|}{63.6}          &\multicolumn{1}{c|}{76.9}    & \textbf{87.5} \\

\Xhline{5\arrayrulewidth}             
\end{tabular}
}
\caption{\textbf{Comparison study on Linemod.} We present the results for ADD(-S) metric and compare them to state of the art. While Linemod is close to saturated for fully supervised methods, one-shot pose estimation is still challenging. PoseMatcher achieves the best results for all objects for the one-shot category, surpassing self-supervised methods and close to fully supervised methods.
Best results for one-shot are bolded. $^*$ denotes symmetric objects.}
\label{tab:linemod}
\end{table*}

We model the $\textit{zoom}$ as a classification problem and discrete the possible zooms in $K$ classes~\cite{SC6D}. 
We output the estimated \textit{zoom} $\delta_{z}$ as the expectation over all classes and 2d translation $\delta_{2d}$ necessary to align $\zeta^0$ with $\zeta^{GT}$ as following:

\vspace{-10pt}
\begin{equation}
    \begin{cases}
        \zeta^{GT}_z  = \zeta^{GT}_z \cdot \delta_{z} , \\
        p(\zeta^{GT})  = p(\zeta^{GT}) ~ \cdot ~ \delta_{2d}, \\ 
    \end{cases} 
\end{equation}

where $p(P) = \mathcal{K}\zeta[0,0,0]^T$ is the 2D projection location of the object's centroid adjusted for the crop bounding box and $K$ are the camera intrinsic parameters.

We supervise the \textit{zoom} and 2D location offset directly with a sparse loss:
\vspace{-10pt}
\begin{equation}
    \begin{cases}
        \mathcal{L}_{z}  = || \delta_{z} - \varepsilon_z ||_1 \\
        \mathcal{L}_{2d}  = || \delta_{2d} - \varepsilon_{2d} ||_1 \\
    \end{cases} 
\end{equation}
where $\varepsilon$ is noise introduced at training time. 

The full loss with our improvements becomes:

\vspace{-10pt}
\begin{equation}
    \mathcal{L} = \mathcal{L}_c + \mathcal{L}_f + \alpha~(\mathcal{L}_{z} + \mathcal{L}_{2d}) .
\end{equation}

\section{Experiments.}

\subsection{Data preparation} 

In order to train PoseMatcher from scratch we must use a sufficiently large dataset to encompass a large number of shapes and textures. We use the Google Scanned Objects\cite{goog_objects} dataset with 1023 household objects. We use the renderings provided by \cite{gen6d} for fair comparison where each object is rendered at 250 different viewpoints. We use an additional set of 500 keypoints from ShapeNet \cite{shapenet} with renderings provided by \cite{6dwild}.
For each instance we find a close viewpoint, $\mathcal{T}^+$,  sampled within $[5^{\circ}, 25^{\circ}]$ orientation to ensure high co-visibility while being sufficiently disparate. $\mathcal{T}^-$ is sampled randomly from a set of the 5 farthest viewpoints from the query, which ensures enough data variety. We apply the standard color and noise augmentations~\cite{surfemb,gdr,crt6d} as well as bounding box \textit{zoom-ins} as proposed by CDPN~\cite{cdpn}. We also perform object level augmentations by changing the object's canonical reference frame. Although only 1500 objects are used, by providing strong augmentations we are able to avoid overfitting.

\subsection{Implementation Details.}

For fair comparison with OnePose++, we use 3 IO-Layers. We set the contrastive temperature term $\tau=0.1$. For 3D-refinement we set $K=100$ and $\alpha=100$.
During training we sample $M=2048$ from each templates for a total of $4096$. For refinement, we use the top 512 matches with confidences up to $\theta=0.1$. If not enough matches during training, we pad the pointcloud with groundtruth matches. At test time, we sample 16k keypoints from the available training images. We prune $50\%$ of the point cloud after each IO-Layer, to 8k and 4k after the first two layers, respectively.

For all experiments, we use AdamW~\cite{adam} with cosine learning rate decay and we linearly warmup the learning rate for 5 epochs. We train PoseMatcher for 50 epochs with a batch size of 8 and an intial learning rate of 0.0001. At each epoch, we sample 10 different viewpoint queries for each of the 1500 objects. 

\subsection{RGB vs Grayscale images}

Interest point descriptor models are mostly used over grayscale images as it has been shown to improve generalization. OnePose and OnePose++~\cite{onepose, onepose++} are applied to grayscale images because they make use of pre-trained descriptor models trained solely on grayscale, SuperPoint~\cite{superpoint} and LoFTR respectively~\cite{loftr}. PoseMatcher does not rely on prior methods thus is not limited to grayscale. To provide a comprehensive analysis and ensure thoroughness we train our method on both RGB and grayscale input. Intuitively, RGB inputs should provide more information and allow PoseMatcher to better separate textured regions of an object as well as from the background. Surprisingly, we found that using RGB data improves our method. Grayscale only reaches an ADD-(S) accuracy of $84.1\%$ on Linemod while RGB inputs reach an accuracy $87.5\%$. All our experiments and further ablations use RGB inputs.

\subsection{Evaluation Results}

Our results on Linemod and the YCB-V datasets use the standard 2D bounding boxes provided by the BOP Challenge~\cite{bop}. To note these are trained using synthetic versions of the target datasets which is done in order for a fair comparison to other methods that rely on the same detections.
 To measure the performance of PoseMatcher, we employ the same metrics used by prior methods. The ADD(-S) metric considers a pose correct if the average distance to the groundtruth falls below a threshold, usually $10\%$ of the objects diameter, with a slightly modified version for symmetric objects~\cite{posecnn}. We also measure the accuracy of the predicted translation under a range of threshold to support our 3D-refinement step.


We outperform every existing one-shot method by a significant margin, specially on more complicated objects such as the Ape or the Cat. We have a large advantage over DeepIM~\cite{deepim} which needs a strong pose initialization from PoseCNN~\cite{posecnn} whereas PoseMatcher does not require any type of pose initialization or the 3D mesh model. It is interesting to observe that we also outperform PVNet\cite{pvnet}, an instance-level pose estimator. 

\subsection{Ablation Studies}

\begin{table}[t]
\resizebox{\columnwidth}{!}{%
\begin{tabular}{cccccc|cc}
\Xhline{5\arrayrulewidth}
Three-View & IO-Layer & Pruning & 2D-Ref & 3D-Ref & Grayscale & Linemod  & YCB-V \\ \midrule
    &       &          &    \checkmark    &        &    &    76.9         &       -         \\ \hline
 \checkmark    &       &          &        &        &  &     81.1          &       22.1         \\ 
\checkmark      &            &          & \checkmark       &        &  &  81.4             &  22.5       \\ \hline
\checkmark  &       \checkmark         &          &        &        &    &  83.4          &    24.1           \\ 
\checkmark  &        \checkmark        &          &   \checkmark     &    &    &   83.4             &   24.2       \\ 
\checkmark  &        \checkmark        &  \checkmark        & \checkmark       &    &    &    83.9    &  25.8   \\ 
\checkmark  &        \checkmark        &  \checkmark        &        & \checkmark   &    &   \textbf{87.5}     &  \textbf{31.3} \\ 
\checkmark  &        \checkmark        &  \checkmark        &        & \checkmark   &  \checkmark  &   84.1     & 27.4 \\ 
\Xhline{5\arrayrulewidth}             
\end{tabular}
} 
\caption{\textbf{Ablation studies for each component.} We present the results on Linemod and YCB-V measured by ADD-(S). The first line are the results gathered from OnePose++~\cite{onepose} The model used for the second and third rows refer is identical to the OnePose++~\cite{onepose++} model trained with our three-view based pipeline.}
\end{table}

Our new training pipeline elevates the results from OnePose++~\cite{onepose++}. By learning meaningful descriptors specifically trained for pose estimation instead of pretrained ones, we are able increase the quality of the feature matching thereby improving the pose accuracy by $5\%$. This confirms that carefully designing a training pipeline can yield better results.

The IO-Layer, our novel attention based layer, yields an improvement of $2\%$ on both datasets. While it does not increase runtime performance when compared to OnePose++~\cite{onepose}, we can see an improvement stemming from the use of specialized modality weights.

\begin{figure}[bp]
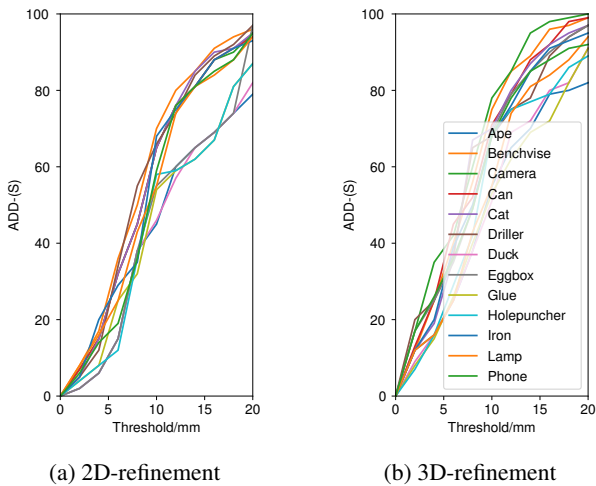
%
  \vspace{-15pt}
    \centering
    \subfloat[\centering 2D-refinement]{{\resizebox{0.44\columnwidth}{!}{
      \input{resources/before_ref.pgf}
      } }}%
    \qquad
    \subfloat[\centering 3D-refinement]{{\resizebox{0.44\columnwidth}{!}{
      \input{resources/after_ref.pgf}
      } }}%
    \caption{\textbf{Difference between refinement operations.} We can see that for all objects in Linemod, the 3D-refinement has a big impact on lower thresholds.}%
    \label{fig:example}%
\vspace{-6pt}
\end{figure}

Replacing 2D based refinement with pose specific 3D-Refinement improved our results significantly. We found improvements in both Linemod and the YCB-V datasets.  The improvement cannot be solely observed by measuring the ADD-(S) since the average threshold for Linemod is $15cm$. We provide further information for Linemod on Fig.~\ref{fig:example} where the error curves for each object in Linemod can be seen. The 3D-refinement has a much larger impact on lower thresholds accuracies than higher ones. An addition like 3D-refinement is invaluable for applications that require high precision.

\begin{figure}[b]
\vspace{-10pt}
    \resizebox{0.9\columnwidth}{!}{
      \input{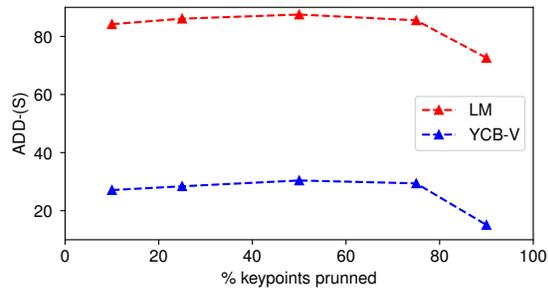}
      }
\caption{\textbf{Pruning ablations.} With this ablation, we show that \Ours~is not reliant on a large number of keypoints. Although significant decreases in performance occur when using 8 keypoints, we see small differences for larger amounts, with the optimal number of keypoints being 64 for both datasets.} 
  \label{fig:pruning}
\end{figure}

We found 2D-refinement to result in no significant improvements. This step does not correct significant matching errors and PnP seems to overcome small imperfections stemming from resolution errors. PnP is also subject to biasing for translation as observed by prior works~\cite{gdr, cdpn}. Our results on 3D-refinement show a large increase in the translation accuracy when we use 3D-refinement over 2D-refinement, which is also reflected on ADD-(S). 


\vspace{-5pt}
\subsubsection{Object pruning ablations}.\label{sec:experiments}
\vspace{-5pt}

As discussed before, pruning an object provides two advantages. Firstly, it eliminates self-occluded regions of the object which removes possible noise during feature matching. Secondly, it reduces the complexity of the attention mechanism by decreasing the amount of keypoints needed to be \textit{attented}, which leads to faster runtime. However, pruning a large amount of keypoints leads to poorer results as it excludes essential keypoints.  We perform inference time ablation studies to help us determine the optimal pruning percentage and plot the results of percentages ranging from $10\%$ to $90\%$ after each IO-Layer in Fig~\ref{fig:pruning} and measure the ADD-(S) accuracy on Linemod. 
If a low amount of keypoints are removed, the performance is not highly affected. However, we see a small drop at $75\%$ with a large drop when crop $90\%$ of the existing pointcloud. The latter might be due to the number of initial keypoints as the number of keypoints are just 150 points.
Although increasing the number of pruned keypoints leads to higher throughput, the peak accuracy results is at $50\%$.

\subsubsection{Number of Templates}

For all our experiments on Linemod, we have been using all the available training images in order to sample template keypoints. Due to our training pipeline, PoseMatcher is robust to incomplete and noisy templates. We perform an ablation study over the number of templates needed. Since each object in Linemod contains different number of training images, we present the percentage of training images used. To sample which images to use, we sample from the available viewpoints using furthest point sampling~\cite{pvnet} w.r.t. orientations in order to cover the widest range of viewpoints. \Ours~almost achieves the same level of accuracy as OnePose++~\cite{onepose++} using only $75\%$ of the available training images.

\begin{table}[]
\resizebox{\columnwidth}{!}{%
\begin{tabular}{c|ccccccccccccc|c}
\Xhline{5\arrayrulewidth}
$\%$  & Ape & BW & Camera & Can & Cat &Driller&Duck&EB*&Glue*&HP&Iron&Lamp&Phone & Avg \\ \midrule
\multicolumn{1}{c|}{10}  &2.1 &15.1& 8.8 &0.0  &1.9 &6.4   &0    &25    &2.5  &0    &12.4&2.1  &12.4 &6.8   \\ 
\multicolumn{1}{c|}{25}  &5.9 &24.1& 15.2&12.4 &13.1&12.4  &1.2  &45.1  &8.3  &0    &24.4&12.6 &29.1 &15.7   \\ 
\multicolumn{1}{c|}{50}  &12.4&59.4& 35.6&19.7 &19.3&34.7  &9.4  &78.9 &41.9 &5.7  &34.1&21.8 &57.3 &33.1   \\ 
\multicolumn{1}{c|}{75}  &39.9&78.9& 81.2&76.6 &57.3&65.1  &27.1 &92.1  &72.1 &46.8 &87.1&79.  &88.5 &68.6   \\ 
\hline
\multicolumn{1}{c|}{100} &59.2&98.1& 93.4&96 &88.0&98.4  &54.1   &97.8  &91.5 &73.4 &97.9&98.1 &92.1 &87.5   \\ 
\Xhline{5\arrayrulewidth}             
\end{tabular}
} 
\caption{\textbf{Number of templates.} We present the results on Linemod for $\%$ of training images used. Each object has around 180 images therefore $10\%$ is only 18 templates to use.}
\vspace{-6pt}
\end{table}

\begin{figure}[t]
    \resizebox{\columnwidth}{!}{%
    \centering 
    \begin{subfigure}{\textwidth}
      \fbox{\includegraphics[width=\linewidth]{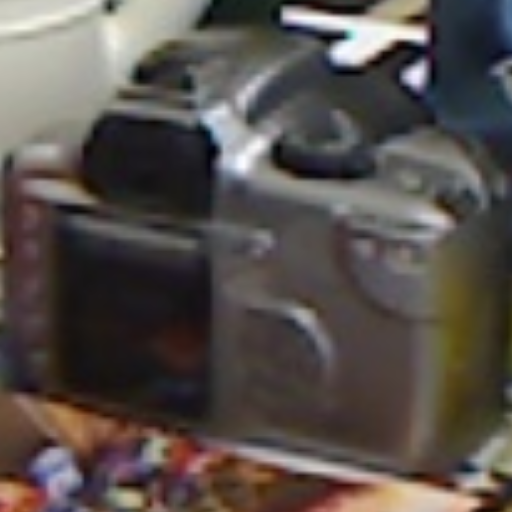}}
      \label{}
    \end{subfigure}\hfil 
    \begin{subfigure}{\textwidth}
      \fbox{\includegraphics[width=\linewidth]{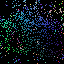}}
      \label{}
    \end{subfigure}\hfil 
    \begin{subfigure}{\textwidth}
      \fbox{\includegraphics[width=\linewidth]{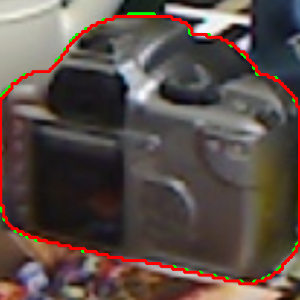}}
      \label{}
    \end{subfigure}
    }
    \vspace{-6pt}
    \medskip 
    \resizebox{\columnwidth}{!}{%
    \begin{subfigure}{\textwidth}
      \fbox{\includegraphics[width=\linewidth]{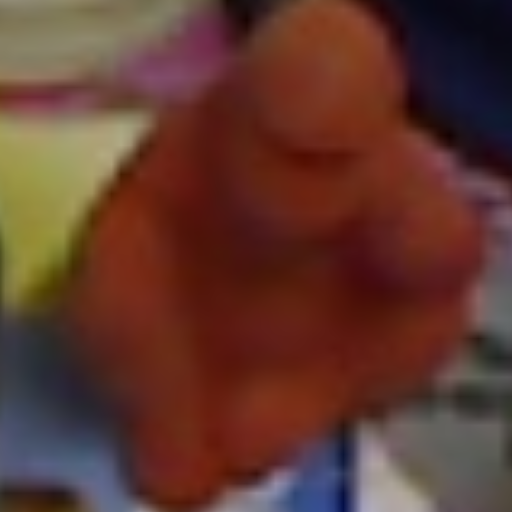}}
      \label{fig:4}
    \end{subfigure}\hfil 
    \begin{subfigure}{\textwidth}
      \fbox{\includegraphics[width=\linewidth]{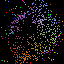}}
      \label{fig:5}
    \end{subfigure}\hfil 
    \begin{subfigure}{\textwidth}
      \fbox{\includegraphics[width=\linewidth]{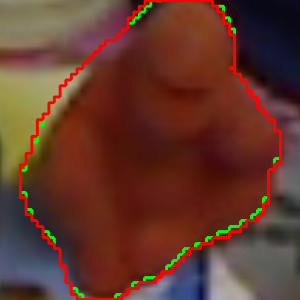}}
      \label{fig:6}
    \end{subfigure}
    }
    \resizebox{\columnwidth}{!}{%
    \begin{subfigure}{\textwidth}
      \fbox{\includegraphics[width=\linewidth]{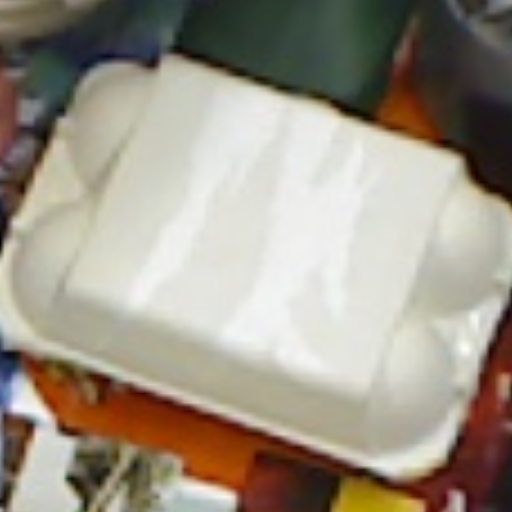}}
      \label{fig:4}
    \end{subfigure}\hfil 
    \begin{subfigure}{\textwidth}
      \fbox{\includegraphics[width=\linewidth]{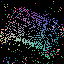}}
      \label{fig:5}
    \end{subfigure}\hfil 
    \begin{subfigure}{\textwidth}
      \fbox{\includegraphics[width=\linewidth]{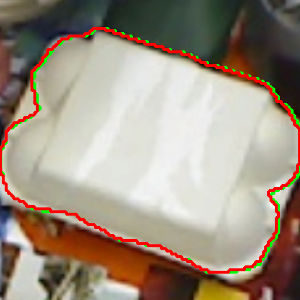}}
      \label{fig:6}
    \end{subfigure}
    }
    \resizebox{\columnwidth}{!}{%
    \begin{subfigure}{\textwidth}
      \fbox{\includegraphics[width=\linewidth]{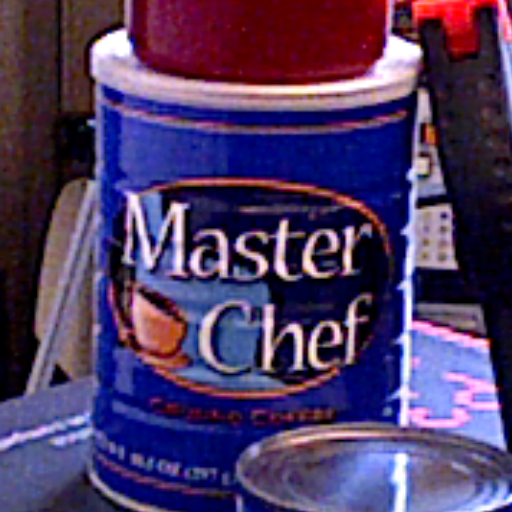}}
      \label{fig:4}
    \end{subfigure}\hfil 
    \begin{subfigure}{\textwidth}
      \fbox{\includegraphics[width=\linewidth]{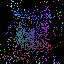}}
      \label{fig:5}
    \end{subfigure}\hfil 
    \begin{subfigure}{\textwidth}
      \fbox{\includegraphics[width=\linewidth]{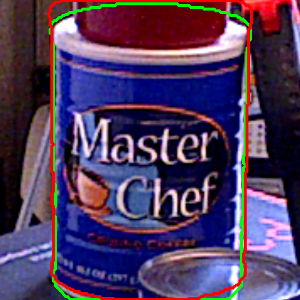}}
      \label{fig:6}
    \end{subfigure}
    }
    \resizebox{\columnwidth}{!}{%
    \begin{subfigure}{\textwidth}
      \fbox{\includegraphics[width=\linewidth]{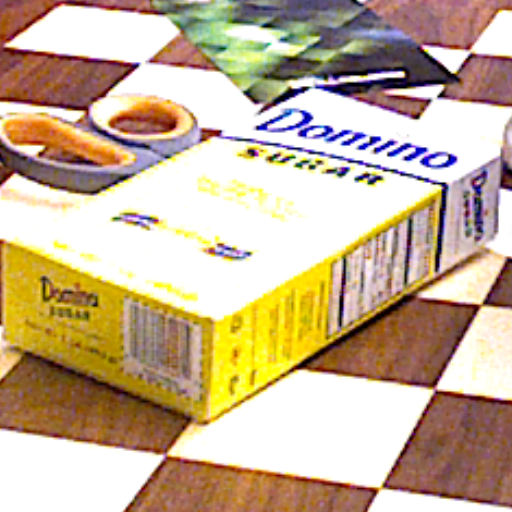}}
      \label{fig:4}
    \end{subfigure}\hfil 
    \begin{subfigure}{\textwidth}
      \fbox{\includegraphics[width=\linewidth]{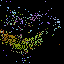}}
      \label{fig:5}
    \end{subfigure}\hfil 
    \begin{subfigure}{\textwidth}
      \fbox{\includegraphics[width=\linewidth]{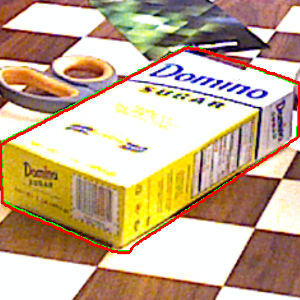}}
      \label{fig:6}
    \end{subfigure}
    }
\vspace{-15pt}
\caption{\textbf{Visualization of matching keypoints.} Here we show PoseMatcher output $\mathcal{C}$, where we assigned a normalized coordinate of the highest confidence keypoint to the corresponding pixel. The last two rows are examples from the YCB-V dataset.}
  \vspace{-15pt}
\label{fig:attention}

\end{figure}
\section{Conclusions}

We proposed \Ours, a novel model free one-shot pose estimator based on deep feature matching. Given a sequence of template images, we can reconstruct a feature point cloud and extract matches from a query image.  In order to avoid using pre-trained descriptor models, we introduce a new training pipeline that allows us to train from scratch. With this simple addition, we show that we can improve OnePose++~\cite{onepose++} without any additional changes. We build on top of OnePose++ by designing a new attention layer IO-Layer, designed specifically for image to object matching. Additionally, we propose improvements to the pipeline including object pruning and 3D based refinement.

\par \textbf{Limitations.} Unfortunately we found that MatchPose has limited domain adaptation capability. If there is a large domain gap between the template domain (ex. synthetic renderings) and queries MatchPose is unable to correctly match keypoints, even coarsely. We specifically find this on YCB-V where each scene contains different levels of sensor noise and illumination. We also found symmetric objects particularly difficult, even though object pruning did show improvements.

{\small
\bibliographystyle{ieee_fullname}
\bibliography{paper}

\begin{thebibliography}{10}\itemsep=-1pt

\bibitem{densebody}
R{\i}za Alp~G{\"u}ler, Natalia Neverova, and Iasonas Kokkinos.
\newblock {DensePose}: Dense human pose estimation in the wild.
\newblock In {\em CVPR}, 2018.

\bibitem{SC6D}
Dingding Cai, Janne Heikkil{\"a}, and Esa Rahtu.
\newblock Sc6d: Symmetry-agnostic and correspondence-free 6d object pose
  estimation.
\newblock {\em arXiv preprint arXiv:2208.02129}, 2022.

\bibitem{detr}
Nicolas Carion, Francisco Massa, Gabriel Synnaeve, Nicolas Usunier, Alexander
  Kirillov, and Sergey Zagoruyko.
\newblock End-to-end object detection with transformers.
\newblock In {\em European conference on computer vision}, pages 213--229.
  Springer, 2020.

\bibitem{crt6d}
Pedro Castro and Tae-Kyun Kim.
\newblock Crt-6d: Fast 6d object pose estimation with cascaded refinement
  transformers.
\newblock In {\em Proceedings of the IEEE/CVF Winter Conference on Applications
  of Computer Vision}, pages 5746--5755, 2023.

\bibitem{shapenet}
Angel~X. Chang, Thomas Funkhouser, Leonidas Guibas, Pat Hanrahan, Qixing Huang,
  Zimo Li, Silvio Savarese, Manolis Savva, Shuran Song, Hao Su, Jianxiong Xiao,
  Li Yi, and Fisher Yu.
\newblock {ShapeNet: An Information-Rich 3D Model Repository}.
\newblock Technical Report arXiv:1512.03012 [cs.GR], Stanford University ---
  Princeton University --- Toyota Technological Institute at Chicago, 2015.

\bibitem{superpoint}
Daniel DeTone, Tomasz Malisiewicz, and Andrew Rabinovich.
\newblock Superpoint: Self-supervised interest point detection and description.
\newblock In {\em Proceedings of the IEEE conference on computer vision and
  pattern recognition workshops}, 2018.

\bibitem{sopose}
Yan Di, Fabian Manhardt, Gu Wang, Xiangyang Ji, Nassir Navab, and Federico
  Tombari.
\newblock So-pose: Exploiting self-occlusion for direct 6d pose estimation.
\newblock In {\em Proceedings of the IEEE/CVF International Conference on
  Computer Vision (ICCV)}, 2021.

\bibitem{goog_objects}
Laura Downs, Anthony Francis, Nate Koenig, Brandon Kinman, Ryan Hickman, Krista
  Reymann, Thomas~B McHugh, and Vincent Vanhoucke.
\newblock Google scanned objects: A high-quality dataset of 3d scanned
  household items.
\newblock In {\em 2022 International Conference on Robotics and Automation
  (ICRA)}, pages 2553--2560. IEEE, 2022.

\bibitem{6dwild}
Yang Fu and Xiaolong Wang.
\newblock Category-level 6d object pose estimation in the wild: A
  semi-supervised learning approach and a new dataset.
\newblock {\em arXiv preprint arXiv:2206.15436}, 2022.

\bibitem{surfemb}
Rasmus~Laurvig Haugaard and Anders~Glent Buch.
\newblock Surfemb: Dense and continuous correspondence distributions for object
  pose estimation with learnt surface embeddings.
\newblock In {\em Proceedings of the IEEE/CVF Conference on Computer Vision and
  Pattern Recognition}, pages 6749--6758, 2022.

\bibitem{resnet}
Kaiming He, Xiangyu Zhang, Shaoqing Ren, and Jian Sun.
\newblock Deep residual learning for image recognition.
\newblock In {\em CVPR}, 2016.

\bibitem{onepose++}
Xingyi He, Jiaming Sun, Yuang Wang, Di Huang, Hujun Bao, and Xiaowei Zhou.
\newblock Onepose++: Keypoint-free one-shot object pose estimation without
  {CAD} models.
\newblock In {\em Advances in Neural Information Processing Systems}, 2022.

\bibitem{ppf}
Stefan Hinterstoisser, Vincent Lepetit, Naresh Rajkumar, and Kurt Konolige.
\newblock Going further with point pair features.
\newblock {\em CoRR}, abs/1711.04061, 2017.

\bibitem{bop}
Tom{\'a}{\v{s}} Hoda{\v{n}}, Frank Michel, Eric Brachmann, Wadim Kehl, Anders
  Glent~Buch, Dirk Kraft, Bertram Drost, Joel Vidal, Stephan Ihrke, Xenophon
  Zabulis, Caner Sahin, Fabian Manhardt, Federico Tombari, Tae-Kyun Kim,
  Ji{\v{r}}{\'i} Matas, and Carsten Rother.
\newblock {Bop}: Benchmark for {6d} object pose estimation.
\newblock {\em ECCV}, 2018.

\bibitem{seg_driven}
Yinlin Hu, Joachim Hugonot, Pascal Fua, and Mathieu Salzmann.
\newblock Segmentation-driven 6d object pose estimation.
\newblock In {\em Proceedings of the IEEE/CVF Conference on Computer Vision and
  Pattern Recognition (CVPR)}, 2019.

\bibitem{repose}
Shun Iwase, Xingyu Liu, Rawal Khirodkar, Rio Yokota, and Kris~M. Kitani.
\newblock Repose: Fast 6d object pose refinement via deep texture rendering.
\newblock In {\em Proceedings of the IEEE/CVF International Conference on
  Computer Vision (ICCV)}, 2021.

\bibitem{linear_attention}
Angelos Katharopoulos, Apoorv Vyas, Nikolaos Pappas, and Fran{\c{c}}ois
  Fleuret.
\newblock Transformers are rnns: Fast autoregressive transformers with linear
  attention.
\newblock In {\em International Conference on Machine Learning}, pages
  5156--5165. PMLR, 2020.

\bibitem{adam}
Diederick~P Kingma and Jimmy Ba.
\newblock Adam: A method for stochastic optimization.
\newblock In {\em International Conference on Learning Representations (ICLR)},
  2015.

\bibitem{catastrophic}
James Kirkpatrick, Razvan Pascanu, Neil Rabinowitz, Joel Veness, Guillaume
  Desjardins, Andrei~A Rusu, Kieran Milan, John Quan, Tiago Ramalho, Agnieszka
  Grabska-Barwinska, et~al.
\newblock Overcoming catastrophic forgetting in neural networks.
\newblock {\em Proceedings of the national academy of sciences},
  114(13):3521--3526, 2017.

\bibitem{cosypose}
Yann Labb{\'e}, Justin Carpentier, Mathieu Aubry, and Josef Sivic.
\newblock Cosypose: Consistent multi-view multi-object 6d pose estimation.
\newblock In {\em European Conference on Computer Vision}, 2020.

\bibitem{nerfpose}
Fu Li, Hao Yu, Ivan Shugurov, Benjamin Busam, Shaowu Yang, and Slobodan Ilic.
\newblock Nerf-pose: A first-reconstruct-then-regress approach for
  weakly-supervised 6d object pose estimation.
\newblock {\em arXiv preprint arXiv:2203.04802}, 2022.

\bibitem{deepim}
Yi Li, Gu Wang, Xiangyang Ji, Yu Xiang, and Dieter Fox.
\newblock Deepim: Deep iterative matching for 6d pose estimation.
\newblock In {\em Proceedings of the European Conference on Computer Vision
  (ECCV)}, 2018.

\bibitem{cdpn}
Zhigang Li, Gu Wang, and Xiangyang Ji.
\newblock Cdpn: Coordinates-based disentangled pose network for real-time
  rgb-based 6-dof object pose estimation.
\newblock In {\em ICCV}, 2019.

\bibitem{focalloss}
Tsung-Yi Lin, Priya Goyal, Ross Girshick, Kaiming He, and Piotr Doll{\'a}r.
\newblock Focal loss for dense object detection.
\newblock In {\em Proceedings of the IEEE international conference on computer
  vision}, 2017.

\bibitem{gen6d}
Yuan Liu, Yilin Wen, Sida Peng, Cheng Lin, Xiaoxiao Long, Taku Komura, and
  Wenping Wang.
\newblock Gen6d: Generalizable model-free 6-dof object pose estimation from rgb
  images.
\newblock In {\em Computer Vision--ECCV 2022: 17th European Conference, Tel
  Aviv, Israel, October 23--27, 2022, Proceedings, Part XXXII}, pages 298--315.
  Springer, 2022.

\bibitem{paraformer}
Xiaoyong Lu, Yaping Yan, Bin Kang, and Songlin Du.
\newblock Paraformer: Parallel attention transformer for efficient feature
  matching.
\newblock {\em arXiv preprint arXiv:2303.00941}, 2023.

\bibitem{heatmaps}
Markus Oberweger, Mahdi Rad, and Vincent Lepetit.
\newblock Making deep heatmaps robust to partial occlusions for {3D} object
  pose estimation.
\newblock In {\em ECCV}, 2018.

\bibitem{latentfusion}
Keunhong Park, Arsalan Mousavian, Yu Xiang, and Dieter Fox.
\newblock Latentfusion: End-to-end differentiable reconstruction and rendering
  for unseen object pose estimation.
\newblock In {\em Proceedings of the IEEE/CVF conference on computer vision and
  pattern recognition}, pages 10710--10719, 2020.

\bibitem{pix2pose}
Kiru Park, Timothy Patten, and Markus Vincze.
\newblock Pix2pose: Pix2pose: Pixel-wise coordinate regression of objects for
  6d pose estimation.
\newblock In {\em ICCV}, 2019.

\bibitem{pvnet}
Sida Peng, Yuan Liu, Qixing Huang, Xiaowei Zhou, and Hujun Bao.
\newblock Pvnet: Pixel-wise voting network for 6dof pose estimation.
\newblock In {\em CVPR}, 2019.

\bibitem{unseen1}
Giorgia Pitteri, Aur{\'e}lie Bugeau, Slobodan Ilic, and Vincent Lepetit.
\newblock 3d object detection and pose estimation of unseen objects in color
  images with local surface embeddings.
\newblock In {\em Proceedings of the Asian Conference on Computer Vision},
  2020.

\bibitem{unseen2}
Giorgia Pitteri, Slobodan Ilic, and Vincent Lepetit.
\newblock Cornet: generic 3d corners for 6d pose estimation of new objects
  without retraining.
\newblock In {\em Proceedings of the IEEE/CVF International Conference on
  Computer Vision Workshops}, pages 0--0, 2019.

\bibitem{bb8}
Mahdi Rad and Vincent Lepetit.
\newblock Bb8: A scalable, accurate, robust to partial occlusion method for
  predicting the 3d poses of challenging objects without using depth.
\newblock In {\em ICCV}, 2017.

\bibitem{superglue}
Paul-Edouard Sarlin, Daniel DeTone, Tomasz Malisiewicz, and Andrew Rabinovich.
\newblock Superglue: Learning feature matching with graph neural networks.
\newblock In {\em Proceedings of the IEEE/CVF conference on computer vision and
  pattern recognition}, pages 4938--4947, 2020.

\bibitem{osop}
Ivan Shugurov, Fu Li, Benjamin Busam, and Slobodan Ilic.
\newblock Osop: A multi-stage one shot object pose estimation framework.
\newblock In {\em Proceedings of the IEEE/CVF Conference on Computer Vision and
  Pattern Recognition}, pages 6835--6844, 2022.

\bibitem{juil}
Juil Sock, Guillermo Garcia-Hernando, Anil Armagan, and Tae-Kyun Kim.
\newblock Introducing pose consistency and warp-alignment for self-supervised
  6d object pose estimation in color images.
\newblock In {\em 2020 International Conference on 3D Vision (3DV)}, pages
  291--300, 2020.

\bibitem{loftr}
Jiaming Sun, Zehong Shen, Yuang Wang, Hujun Bao, and Xiaowei Zhou.
\newblock {LoFTR}: Detector-free local feature matching with transformers.
\newblock {\em CVPR}, 2021.

\bibitem{onepose}
Jiaming Sun, Zihao Wang, Siyu Zhang, Xingyi He, Hongcheng Zhao, Guofeng Zhang,
  and Xiaowei Zhou.
\newblock {OnePose}: One-shot object pose estimation without {CAD} models.
\newblock {\em CVPR}, 2022.

\bibitem{seamless}
Bugra Tekin, Sudipta~N Sinha, and Pascal Fua.
\newblock {Real-Time} seamless single shot {6D} object pose prediction.
\newblock In {\em CVPR}, 2018.

\bibitem{dualsoftmax}
Micha{\l} Tyszkiewicz, Pascal Fua, and Eduard Trulls.
\newblock Disk: Learning local features with policy gradient.
\newblock {\em Advances in Neural Information Processing Systems},
  33:14254--14265, 2020.

\bibitem{self6d}
Gu Wang, Fabian Manhardt, Jianzhun Shao, Xiangyang Ji, Nassir Navab, and
  Federico Tombari.
\newblock {Self6D}: Self-supervised monocular 6d object pose estimation.
\newblock In {\em European Conference on Computer Vision (ECCV)}, pages
  108--125, 2020.

\bibitem{gdr}
Gu Wang, Fabian Manhardt, Federico Tombari, and Xiangyang Ji.
\newblock Gdr-net: Geometry-guided direct regression network for monocular 6d
  object pose estimation.
\newblock In {\em Proceedings of the IEEE/CVF Conference on Computer Vision and
  Pattern Recognition (CVPR)}, 2021.

\bibitem{nocs}
He Wang, Srinath Sridhar, Jingwei Huang, Julien Valentin, Shuran Song, and
  Leonidas~J. Guibas.
\newblock Normalized object coordinate space for category-level 6d object pose
  and size estimation.
\newblock In {\em Proceedings of the IEEE/CVF Conference on Computer Vision and
  Pattern Recognition (CVPR)}, June 2019.

\bibitem{matchformer}
Qing Wang, Jiaming Zhang, Kailun Yang, Kunyu Peng, and Rainer Stiefelhagen.
\newblock Matchformer: Interleaving attention in transformers for feature
  matching.
\newblock In {\em Asian Conference on Computer Vision}, 2022.

\bibitem{posecnn}
Yu Xiang, Tanner Schmidt, Venkatraman Narayanan, and Dieter Fox.
\newblock {PoseCNN}: A convolutional neural network for {6D} object pose
  estimation in cluttered scenes.
\newblock {\em Robotics: Science and Systems (RSS)}, 2018.

\bibitem{posefromshape}
Yang Xiao, Xuchong Qiu, Pierre-Alain Langlois, Mathieu Aubry, and Renaud
  Marlet.
\newblock Pose from shape: Deep pose estimation for arbitrary 3d objects.
\newblock {\em arXiv preprint arXiv:1906.05105}, 2019.

\bibitem{inerf}
Lin Yen-Chen, Pete Florence, Jonathan~T Barron, Alberto Rodriguez, Phillip
  Isola, and Tsung-Yi Lin.
\newblock inerf: Inverting neural radiance fields for pose estimation.
\newblock In {\em 2021 IEEE/RSJ International Conference on Intelligent Robots
  and Systems (IROS)}, pages 1323--1330. IEEE, 2021.

\bibitem{dpod}
Sergey Zakharov, Ivan Shugurov, and Slobodan Ilic.
\newblock Dpod: 6d pose object detector and refiner.
\newblock In {\em ICCV}, 2019.

\bibitem{wild6d2}
Kaifeng Zhang, Yang Fu, Shubhankar Borse, Hong Cai, Fatih Porikli, and Xiaolong
  Wang.
\newblock Self-supervised geometric correspondence for category-level 6d object
  pose estimation in the wild.
\newblock {\em arXiv preprint arXiv:2210.07199}, 2022.

\end{thebibliography}
}

\end{document}